# Bounded Planning in Passive POMDPs

**Roy Fox** ROYF@CS.HUJI.AC.IL
**Naftali Tishby** TISHBY@CS.HUJI.AC.IL
School of Computer Science and Engineering
The Hebrew University
Jerusalem 91904, Israel

## Abstract

In Passive POMDPs actions do not affect the world state, but still incur costs. When the agent is bounded by information-processing constraints, it can only keep an approximation of the belief. We present a variational principle for the problem of maintaining the information which is most useful for minimizing the cost, and introduce an efficient and simple algorithm for finding an optimum.

## 1. Introduction

### 1.1. Passive POMDPs Planning

Planning in Partially Observable Markov Decision Processes (POMDPs) is an important task in reinforcement learning, which models an agent's interaction with its environment as a discrete-time stochastic process. The environment goes through a sequence of *world states* $W_1, \ldots, W_n$ in a finite domain $\mathcal{W}$. These states are hidden from the agent except for an observation $O_t$ in a finite domain $\mathcal{O}$, distributed by $\sigma(O_t|W_t)$.

In the standard POMDP, the agent then chooses an action, which affects the next world state and incurs a cost. Here we consider *Passive POMDPs*, in which the action affects the cost, but not the world state. We assume that the world itself is a Markov Chain, with states governed by a time-independent transition probability function $p(W_t|W_{t-1})$ and an initial distribution $P_1(W_1)$.

The agent maintains an internal memory state $M_t$ in a finite domain $\mathcal{M}$. In each step the memory state is updated from the previous memory state and the current observation, according to a memory-state transition function $q_t(M_t|M_{t-1}, O_t)$ which serves as an *inference policy*. Figure 1 summarizes the stochastic process.

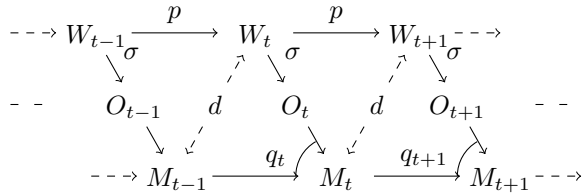

*Figure 1.* Structure of the Bayes network model of Passive POMDP planning

The agent's goal is to minimize the average expected cost of its actions. In this paper we take the agent's memory state to represent the action, and define a cost function $d : \mathcal{W} \times \mathcal{M} \to \mathbb{R}$ on the world and memory states. The planning task is then to minimize

$$\frac{1}{n} \sum_{t=1}^{n} \mathop{\mathrm{E}}_{W_t, M_t} d(W_t, M_t)$$

given the model parameters $P_1$, $p$, $\sigma$ and $d$.

A Passive POMDP can be viewed as an HMM in which inference quality is measured by a cost function. Examples of Passive POMDPs include various gambling scenarios, such as the stock exchange or horse racing, where the betting does not affect the world state. In some settings, the reward depends directly on the amount of information that the agent has on the world state (Kelly gambling, see Cover & Thomas, 2006).

When the agent is unbounded it has a simple deterministic optimal inference policy. It can maintain a *belief* $B_t(W_t|O_{(t)})$, which is the posterior probability of the world state $W_t$ given the observable history $O_{(t)} = O_1, \ldots, O_t$. The belief is a minimal sufficient statistic of $O_{(t)}$ for $W_t$, and therefore keeps all the relevant information. It can be computed sequentially by a forward algorithm, starting with $B_1(W_1|O_1) \propto P_1(W_1)\sigma(O_1|W_1)$, and at each step updating

$$B_t(W_t|O_{(t)})$$
$$\propto \sum_{w_{t-1}} B_{t-1}(w_{t-1}|O_{(t-1)})p(W_t|w_{t-1})\sigma(O_t|W_t),$$

normalized to be a probability vector.

## 1.2. Information Constraints

The sufficiency of the exact belief allows the agent to minimize the external cost, but it incurs significant internal costs. The amount of information which the agent needs to keep in memory can be large, and even each observation can be more than the agent can grasp. Anyway, not all of this information is equally useful in reducing external costs.

In general, the agent's information-processing capacity may be bounded in two ways:

1. The capacity of the agent's memory may limit its information rate between $M_{t-1}$ and $M_t$, to $R_M$.
2. The capacity of the channel from the agent's sensors to its memory may limit the rate at which the agent is able to process the observation $O_t$ while it is available, to $R_S$.

The requirement that the agent keeps sufficient statistics and exact beliefs is unrealistic. Rather, the agent's memory $M_t$ must be a statistic of $O_{(t)}$ which is not sufficient, but is still "good" in the sense that it keeps the external cost low. We also want it to be "minimal" for that level of quality, in terms of information-processing rates, so that the agent keeps only information which is useful enough. For each step individually, this is exactly the notion captured by rate-distortion theory, and here we have a sequential extension of it.

The main results of this paper are the formulation of the setting described above, and the introduction of an efficient and simple algorithm to solve it. We prove that the algorithm converges to a local optimum, and demonstrate in simulations the tradeoff of memory and sensing intrinsic to this setting. The application of our results to previously studied problems, and a comparison to existing algorithms, are left for future work.

This paper is organized as follows. In section 2 we formulate out setting in information-theoretical terms. In section 3 we solve the problem for one step by finding a variational principle and an efficient optimization algorithm. In section 4 we analyze the complete sequential problem and introduce an algorithm to solve it. In section 5 we show two simulations of our solution.

## 1.3. Related Work

Unconstrained planning in Passive POMDPs is easily done by maintaining the exact belief, and choosing each action to minimize the subjective expected cost. Planning in general POMDPs is harder, in one aspect due to the size of the belief space. Many algorithms plan efficiently but approximately by focusing on a subset of this space.

Several works do so by optimizing a finite-state controller of a given size (Poupart & Boutilier, 2003; Amato et al., 2010). The belief represented by each state of the controller is then the posterior probability of the world state given that memory state. A different approach is to explicitly select a subset of beliefs, and use them to guide the iterations (Pineau et al., 2003). Another is to reduce the dimension of the belief space to its principle components (Roy & Gordon, 2002).

In this paper we present the novel setting of planning in Passive POMDPs which is constrained by information capacities. This setting allows treatment of reinforcement learning in an information-theoretic framework. It may also provide a principled method for belief approximation in general POMDPs. With a fixed action policy the POMDP becomes a Passive POMDP, and a bounded inference policy can be computed. This reduces the belief space, which in turn guides the action planning. This decoupling is similar to Chrisman (1992), and will be explored in future work.

Some research treats POMDPs where the cost is the $D_{KL}$ between the distributions of the next world state when it is controlled and uncontrolled (Todorov, 2006; Kappen et al., 2009). This has interesting analogies to our setting. Our information-rate constraints define, in effect, components of the cost which are the $D_{KL}$ between the distribution of the next *memory* state and its marginals (see section 3.1). Tishby & Polani (2011) combine similar information-rate constraints of perception and action together. Future work will explore and exploit this symmetry in the special case where the memory information rate is unconstrained.

## 2. Preliminaries

Assume that the model parameters $P_1$, $p$, $\sigma$ and $d$ are given. The agent strives to find an inference policy $q_{(n)}$ such that the average expected cost satisfies

$$\frac{1}{n}\sum_{t=1}^{n} \mathop{\mathrm{E}}_{W_t, M_t} d(W_t, M_t) \leq D.$$

for the minimal $D$ possible. However, the agent operates under capacity constraints on the channels from $M_{t-1}$ and $O_t$ to $M_t$. The external cost $d$ parallels the distortion in rate-distortion theory, while the internal costs are information rates. The agent actually needs to minimize a combination of these costs.

Note that the agent will generally have some information on the next observation even before seeing it, i.e. $M_{t-1}$ and $O_t$ will not be independent. The agent therefore has some freedom in choosing what part of the information common to $M_{t-1}$ and $O_t$ it remembers, and what part it forgets and observes anew.

The average information rate in both channels combined cannot exceed their total capacity, that is

$$\frac{1}{n}\sum_{t=1}^{n} I(M_{t-1}, O_t; M_t) \leq R_M + R_S.$$

In addition, in each step the portion of the above information that is absent from $O_t$ may only be passed on the memory channel, and so

$$\frac{1}{n}\sum_{t=1}^{n} I(M_{t-1}; M_t|O_t) \leq R_M.$$

Similarly, information absent from $M_{t-1}$ is subject to the sensory channel capacity

$$\frac{1}{n}\sum_{t=1}^{n} I(O_t; M_t|M_{t-1}) \leq R_S.$$

The distortion constraint and the three information-rate constraints together form the problem of inference-planning in Passive POMDPs (*Problem 1*).

The emergence of three information-rate constraints for two channels is similar in spirit to multiterminal source coding (Berger, 1977). In their terminology, the agent needs to implement in each $M_t$ a lossy coding of the correlated sources $M_{t-1}$ and $O_t$, under capacity constraints, so as to minimize an average expected distortion. The main difference is that here we chose to allow the encoding not to be distributed, in keeping with the ability of memory to interact with perception in biological agents (Laeng & Endestad, 2012).

## 3. One-Step Optimization

### 3.1. Variational Principle

Before we consider the long-term planning required of the agent in Problem 1, we focus on the choice of $q_n$ in the final step, given the other transitions, that is, given the joint distribution of $M_{n-1}$, $W_n$ and $O_n$. We define the *joint belief* $\theta_n(M_{n-1}, W_n)$ to be the joint distribution of $M_{n-1}$ and $W_n$, and have

$$\Pr_{\theta_n}(M_{n-1}, W_n, O_n) = \theta_n(M_{n-1}, W_n)\sigma(O_n|W_n).$$

We are interested in the rate-distortion region which includes all points $(R_M, R_S, D)$ which are *achievable*, that is, for which there exists some $q_n(M_n|M_{n-1}, O_n)$ with

$$\mathcal{D}_{\theta_n}(q_n) \stackrel{def}{=} \mathop{\mathrm{E}}_{W_n, M_n} d(W_n, M_n) \leq D$$

$$\mathcal{I}_{C,\theta_n}(q_n) \stackrel{def}{=} I(M_{n-1}, O_n; M_n) \leq R_M + R_S$$

$$\mathcal{I}_{M,\theta_n}(q_n) \stackrel{def}{=} I(M_{n-1}; M_n|O_n) \leq R_M$$

$$\mathcal{I}_{S,\theta_n}(q_n) \stackrel{def}{=} I(O_n; M_n|M_{n-1}) \leq R_S.$$

For any information-rate pair $(R_M, R_S)$, the minimal achievable $D$ lies on the boundary $\mathcal{D}^*_{\theta_n}(R_M, R_S)$ of the rate-distortion region. When $\theta_n$ and $q_n$ are clear from context, we refer to these quantities as $\mathcal{D}$, $\mathcal{I}_C$, $\mathcal{I}_M$, $\mathcal{I}_S$ and $\mathcal{D}^*$. We find $\mathcal{D}^*(R_M, R_S)$ by minimizing the expected distortion under information-rate constraints. The minimum exists because all our formulas are continuous, and the solution space for $q_n$ is closed.

Let $\bar{q}_n(M_n|M_{n-1})$, $\bar{q}_n(M_n|O_n)$ and $\bar{q}_n(M_n)$ be the marginals of $q_n(M_n|M_{n-1}, O_n)$. We expand the terms of the problem using these conditional probability distributions, to have

$$\min_{q_n, \bar{q}_n} \mathop{\mathrm{E}}_{M_{n-1}, W_n, O_n} \sum_{m_n} q_n(m_n|M_{n-1}, O_n)d(W_n, m_n)$$

$$\mathop{\mathrm{E}}_{M_{n-1}, O_n} \mathrm{D}_{\mathrm{KL}}(q_n(M_n|M_{n-1}, O_n); \bar{q}_n(M_n)) \leq R_M + R_S$$

$$\mathop{\mathrm{E}}_{M_{n-1}, O_n} \mathrm{D}_{\mathrm{KL}}(q_n(M_n|M_{n-1}, O_n); \bar{q}_n(M_n|O_n)) \leq R_M$$

$$\mathop{\mathrm{E}}_{M_{n-1}, O_n} \mathrm{D}_{\mathrm{KL}}(q_n(M_n|M_{n-1}, O_n); \bar{q}_n(M_n|M_{n-1})) \leq R_S$$

under normalization constraints.[1] We may waive the constraints of non-negative probabilities, which will essentially never be active as we shall see later. Also note that we optimize over $q_n$ and $\bar{q}_n$ as distinct parameters. This is justified by theorem 1 which states that, at the optimum, $\bar{q}_n$ are indeed the marginals of $q_n$.

Let the Lagrange multipliers for the constraints be $\gamma_C$, $\gamma_M$ and $\gamma_S$, and their sum $\gamma = \gamma_C + \gamma_M + \gamma_S$. Leaving aside terms of $\log q_n$, the pointwise terms in the Lagrangian will be

$$G(d, \bar{q}_n, M_{n-1}, W_n, O_n, M_n)$$

$$= d(W_n, M_n) - \gamma_C \log \bar{q}_n(M_n)$$

$$-\gamma_M \log \bar{q}_n(M_n|O_n) - \gamma_S \log \bar{q}_n(M_n|M_{n-1}).$$

In the following analysis, several expectations of this function will be useful:

- $G_{\theta_n}(d, \bar{q}_n, M_{n-1}, O_n, M_n)$
  $$= \mathop{\mathrm{E}}_{W_n|M_{n-1}, O_n} G(d, \bar{q}_n, M_{n-1}, W_n, O_n, M_n),$$

- $G_{q_n}(d, \bar{q}_n, M_{n-1}, W_n)$
  $$= \mathop{\mathrm{E}}_{O_n, M_n|M_{n-1}, W_n} G(d, \bar{q}_n, M_{n-1}, W_n, O_n, M_n),$$

- $G_{\theta_n, q_n}(d, \bar{q}_n)$
  $$= \mathop{\mathrm{E}}_{M_{n-1}, W_n, O_n, M_n} G(d, \bar{q}_n, M_{n-1}, W_n, O_n, M_n)$$
  $$= \mathcal{D}_{\theta_n}(q_n) + \gamma_C H(\bar{q}_n(M_n))$$
  $$+ \gamma_M H(\bar{q}_n(M_n|O_n)) + \gamma_S H(\bar{q}_n(M_n|M_{n-1})),$$

---

[1] The information-rate constraints result from the $n$-step Problem 1 by fixing the first $n-1$ steps, if we consider that only two of the constraints are actually used in any instance (see corollary 3).

where $H$ is the entropy function. The Lagrangian of the problem, up to normalization terms and additive constants, can now be written as

$$\mathcal{L}_1(q_n, \bar{q}_n; \theta_n, \gamma_C, \gamma_M, \gamma_S) = G_{\theta_n, q_n}(d, \bar{q}_n) - \gamma H(q_n).$$

### 3.2. Properties of the One-Step Lagrangian

**Theorem 1.** *For any fixed $\theta_n$, $\mathcal{L}_1$ is convex in $q_n$ and $\bar{q}_n$. $\mathcal{L}_1$ is strictly convex in parameters which are conditional on $m_{n-1}$ and $o_n$ with $\Pr_{\theta_n}(m_{n-1}, o_n) > 0$, and at the minimum these satisfy*

$$q_n(M_n | M_{n-1}, O_n) \tag{1}$$
$$= \frac{\exp(-\gamma^{-1} G_{\theta_n}(d, \bar{q}_n, M_{n-1}, O_n, M_n))}{Z_n(M_{n-1}, O_n)},$$

*where $Z_n$ is a normalizing partition function, and*

$$\bar{q}_n(M_n) = \sum_{m_{n-1}, o_n} \Pr_{\theta_n}(m_{n-1}, o_n) q_n(M_n | m_{n-1}, o_n)$$
$$\bar{q}_n(M_n | O_n) = \sum_{m_{n-1}} \Pr_{\theta_n}(m_{n-1} | O_n) q_n(M_n | m_{n-1}, O_n)$$
$$\bar{q}_n(M_n | M_{n-1}) = \sum_{o_n} \Pr_{\theta_n}(o_n | M_{n-1}) q_n(M_n | M_{n-1}, o_n).$$
$$\tag{2}$$

*Proof.* For any fixed $\theta_n$, $\mathcal{L}_1$ is convex since all its terms are convex. Non-zero terms only involve $m_{n-1}$ and $o_n$ with $\Pr_{\theta_n}(m_{n-1}, o_n) > 0$. Focusing on these parameters, the distortion terms are linear, and the information terms strictly convex. The unique feasible extremum of $\mathcal{L}_1$ is then the global minimum. Differentiating by each parameter gives equations 1 and 2. $\square$

If follows from theorem 1 that complementary slackness conditions are sufficient for optimality. Table 1 shows these conditions, the information rates $(R_M, R_S)$ where the solution meets the boundary, and a subgradient of the boundary at that point. For example, if the minimum of $\mathcal{L}_1$ with $\gamma_M = \gamma_S = 0$ satisfies $\mathcal{I}_C \geq \mathcal{I}_M + \mathcal{I}_S$, then for any information-rate pair in the interval $[(\mathcal{I}_C - \mathcal{I}_S, \mathcal{I}_S), (\mathcal{I}_M, \mathcal{I}_C - \mathcal{I}_M)]$ the minimal achievable distortion is $\mathcal{D}$ and $(-\gamma_C, -\gamma_C)$ is a subgradient of the boundary.

**Theorem 2.** *For any joint belief $\theta_n$, the boundary $\mathcal{D}^*_{\theta_n}(R_M, R_S)$ of the rate-distortion region is continuous and convex. Any point $(R_M, R_S, D)$ on the boundary at which $(-\alpha_M, -\alpha_S)$ is a subgradient, is achieved by minimizing $\mathcal{L}_1$ for multipliers*

$$(\gamma_C, \gamma_M, \gamma_S) = \begin{cases} (0, \alpha_M, \alpha_S) & \text{if } \mathcal{I}_C \leq \mathcal{I}_M + \mathcal{I}_S \\ (\alpha_M, 0, \alpha_S - \alpha_M) & \text{if } \mathcal{I}_C \geq \mathcal{I}_M + \mathcal{I}_S \\ & \text{and } \alpha_M \leq \alpha_S \\ (\alpha_S, \alpha_M - \alpha_S, 0) & \text{if } \mathcal{I}_C \geq \mathcal{I}_M + \mathcal{I}_S \\ & \text{and } \alpha_M \geq \alpha_S \end{cases}$$

Table 1. Achievability of the rate-distortion boundary by a minimizer of $\mathcal{L}_1$; If the shown *Conditions* are met by the multipliers and the minimum of $\mathcal{L}_1$, then $\mathcal{D}$ is the minimal distortion for the shown *Rates*, and the shown *Subgradient* is a subgradient of $\mathcal{D}^*$ at that point

| Conditions | Rates | Subgradient |
|---|---|---|
| $\gamma_C = 0$ $\mathcal{I}_C \leq \mathcal{I}_M + \mathcal{I}_S$ | $(\mathcal{I}_M, \mathcal{I}_S)$ | $(-\gamma_M, -\gamma_S)$ |
| $\gamma_M = 0$ $\mathcal{I}_C \geq \mathcal{I}_M + \mathcal{I}_S$ | $(\mathcal{I}_C - \mathcal{I}_S, \mathcal{I}_S)$ | $(-\gamma_C, -\gamma_C - \gamma_S)$ |
| $\gamma_S = 0$ $\mathcal{I}_C \geq \mathcal{I}_M + \mathcal{I}_S$ | $(\mathcal{I}_M, \mathcal{I}_C - \mathcal{I}_M)$ | $(-\gamma_C - \gamma_M, -\gamma_C)$ |
| $\gamma_M = \gamma_S = 0$ $\mathcal{I}_C \geq \mathcal{I}_M + \mathcal{I}_S$ | $[(\mathcal{I}_C - \mathcal{I}_S, \mathcal{I}_S),$ $(\mathcal{I}_M, \mathcal{I}_C - \mathcal{I}_M)]$ | $(-\gamma_C, -\gamma_C)$ |

*Proof.* Let transitions $q_n$ and $q'_n$ achieve the rate-distortion boundary at $(R_M, R_S, D)$ and $(R'_M, R'_S, D')$, respectively, and let $0 \leq \lambda \leq 1$. Then by equations 2 and the convexity of the Kullback-Leibler divergence, the transition $\lambda q_n + (1 - \lambda) q'_n$ (over-)achieves the rate-distortion constraints $\lambda(R_M, R_S, D) + (1 - \lambda)(R'_M, R'_S, D')$. The rate-distortion region is therefore convex, and so is its boundary. The boundary is continuous by the continuity of the problem.

For a positive information-rate pair $(R_M, R_S)$, having $M_n$ independent of $M_{n-1}$ and $O_n$ makes all information-rate constraints inactive. This satisfies the Slater condition, and the multipliers detailed in the theorem are then the Karush-Kuhn-Tucker multipliers necessary for $q_n$ to be optimal. $\square$

**Corollary 3.** *Let $\mathcal{D}^*_C$, $\mathcal{D}^*_M$ and $\mathcal{D}^*_S$ be the boundaries of the rate-distortion regions obtained by keeping each two of the three information-rate constraints. Then $\mathcal{D}^*$ is their maximum.*

### 3.3. Optimization Algorithm

An algorithm which alternatingly minimizes $\mathcal{L}_1$ over each parameter with the others fixed, in the style of Blahut-Arimoto (Cover & Thomas, 2006), will allow us to find the minimum.

**Theorem 4.** *Algorithm 1 converges[2] monotonically to the global minimum of $\mathcal{L}_1$.*

*Proof.* $\mathcal{L}_1$ is non-increasing in each iteration and is bounded from below, which guarantees its monotonic convergence. That is

---

[2] For the sake of clarity, here and in the rest of this paper strict convexity, uniqueness of minimum and convergence should all be taken with respect to events and transitions of positive probability, as justified by theorem 1.

**Algorithm 1** Last-Step Optimization

**Input:** $P_1, p, \sigma, d, \gamma_C, \gamma_M, \gamma_S, \theta_n$
**Output:** optimal $q_n$
  $r \leftarrow 0$
  Initialize some suggestion for $q_n^r$
  **repeat**
    Compute the marginals $\bar{q}_n^r$ of $q_n^r$ (eq. 2)
    Compute a new value for $q_n^{r+1}$ from $\bar{q}_n^r$ (eq. 1)
    $r \leftarrow r+1$
  **until** $q_n^r$ converges

$$\mathcal{L}_1(q_n^r, \bar{q}_n^r) - \mathcal{L}_1(q_n^{r+1}, \bar{q}_n^r) \xrightarrow[r \to \infty]{} 0.$$

But $q_n^{r+1}$ is the unique minimum of the continuous Lagrangian. This implies that $q_n^r$ also converges to a solution $q_n^*$ with marginals $\bar{q}_n^*$. By the continuity of the Lagrangian's derivatives, they are all 0 at this solution. □

## 4. Sequential Rate-Distortion

### 4.1. Variational Principle

Returning to the entire process of Problem 1, the sequence of joint beliefs $\theta_{(2,n)} = \theta_2, \ldots, \theta_n$ depends recursively on $\theta_1$ and the policy $q_{(n)}$. For each $1 \leq t < n$

$$\theta_{t+1}(M_t, W_{t+1}) \tag{3}$$

$$= \sum_{m_{t-1}, w_t} \theta_t(m_{t-1}, w_t) \Pr_{q_t}(M_t, W_{t+1} | m_{t-1}, w_t),$$

with $\theta_1$ given as the independent distribution of $M_0$ and $W_1$.

Adding the constraints of equation 3 with multipliers $\nu_{t, m_t, w_{t+1}}$, the Lagrangian of Problem 1 is

$$\mathcal{L}_n(q_{(n)}, \bar{q}_{(n)}, \theta_{(2,n)}) = \frac{1}{n} \sum_{t=1}^n \mathcal{L}_1(q_t, \bar{q}_t; \theta_t, \gamma_C, \gamma_M, \gamma_S)$$

$$-\frac{1}{n} \sum_{t=1}^{n-1} \sum_{m_t, w_{t+1}} \nu_{t, m_t, w_{t+1}} \left( \theta_{t+1}(m_t, w_{t+1}) \right.$$

$$\left. - \sum_{m_{t-1}, w_t} \theta_t(m_{t-1}, w_t) \Pr_{q_t}(m_t, w_{t+1}|m_{t-1}, w_t) \right)$$

up to normalization terms and additive constants.

Solving $\mathcal{L}_n$ is much more difficult than $\mathcal{L}_1$. $\mathcal{L}_n$ is not convex, and each step may affect all future steps. Intuitively, remembering some feature of the sample in one step is less rewarding if this information is discarded in a future step, and vice versa. This leads to $\mathcal{L}_n$ having many local minima.

### 4.2. Local Optimization Algorithm

Nevertheless, Problem 1 still has some structure which can be insightful to explore. In particular, it has some interesting similarities to the standard POMDP planning problem. Differentiating $\mathcal{L}_n$ by $q_t$ we now get

$$q_t(M_t|M_{t-1}, O_t) \tag{4}$$

$$= \frac{\exp(-\gamma^{-1} G_{\theta_t}(d_t^{\vec{\nu}_t}, \bar{q}_t, M_{t-1}, O_t, M_t))}{Z_t(M_{t-1}, O_t)},$$

with

$$d_t^{\vec{\nu}_t}(W_t, M_t) = d(W_t, M_t) + \mathop{\mathrm{E}}_{W_{t+1}|W_t} \nu_{t, M_t, W_{t+1}},$$

where $\vec{\nu}_n = 0$. $q_t$ now depends on the future through the multiplier vector $\vec{\nu}_t$. Note how the expectation of $\nu_{t, M_t, W_{t+1}}$ given $W_t$ plays a parallel role to that of $d(W_t, M_t)$.

$\mathcal{L}_n$ is linear in each $\theta_t$, and at the optimum must in fact be constant in every non-trivial component of $\theta_t$. This gives us a recursive formula for computing $\vec{\nu}_{t-1}$ from $\vec{\nu}_t$, $q_t$ and $\bar{q}_t$. For $1 < t \leq n$, and whenever $0 < \theta_t(M_{t-1}, W_t) < 1$, we have

$$\nu_{t-1, M_{t-1}, W_t} = G_{q_t}(d_t^{\vec{\nu}_t}, \bar{q}_t, M_{t-1}, W_t) \tag{5}$$

$$- \gamma \mathop{\mathrm{E}}_{O_t|W_t} H(q_t(M_t|M_{t-1}, O_t)) + \lambda_{t, W_t}.$$

Note that equation 5 is a linear backward recursion for $\vec{\nu}_t$. The multipliers $\vec{\lambda}_t$ come from the constraints that $\theta_t$ is a probability distribution function. It has no consequence, however, since it is independent of $M_{t-1}$, and is normalized out when $\vec{\nu}_{t-1}$ is used to compute $q_{t-1}$ in equation 4.

At this point, we can introduce the following generalization of algorithm 1, which finds the optimal transition $q_t$, given the joint belief $\theta_t$ and the policy suffix $q_{(t+1,n)} = q_{t+1}, \ldots, q_n$.

**Algorithm 2** One-Step Optimization

**Input:** $P_1, p, \sigma, d, \gamma_C, \gamma_M, \gamma_S, \theta_t, q_{(t+1,n)}$
**Output:** optimal $q_t$
  $r \leftarrow 0$
  Initialize some suggestion for $q_t^r$
  **repeat**
    Compute $\theta_{(t+1,n)}^r$ from $\theta_t$ and $q_{(t,n-1)}^r$ (eq. 3)
    Compute the marginals $\bar{q}_{(t,n)}^r$ of $q_{(t,n)}^r$ (eq. 2)
    Compute $\vec{\nu}_{(t,n-1)}^r$ recursively backward (eq. 5)
    Compute $q_t^{r+1}$ from $\theta_t^r$, $\bar{q}_t^r$ and $\vec{\nu}_t^r$ (eq. 4)
    $r \leftarrow r+1$
  **until** $q_t^r$ converges

This is a forward-backward algorithm. In each iteration we compute $\theta_{(t+1,n)} = \theta_{t+1}, \ldots, \theta_n$ recursively forward, and then $\vec{\nu}_{(t,n-1)} = \vec{\nu}_t, \ldots, \vec{\nu}_{n-1}$ recursively backward. The algorithm is guaranteed to converge monotonically to an optimal solution, since $\mathcal{L}_n$ is still strictly convex in each $q_t$ separately. In fact, all our theorems and proofs regarding algorithm 1 carry over to this generalization.

### 4.3. Joint-Belief MDP

Expanding the recursion of $\vec{\nu}_t$ in equation 5 to a closed form, and disregarding $\vec{\lambda}_t$, we find that for $1 < t \leq n$ and consistent parameters[3]

$$\mathcal{L}_{n-t+1}(q_{(t,n)}; \theta_t) \qquad (6)$$
$$= \frac{1}{n-t+1} \sum_{m_{t-1}, w_t} \theta_t(m_{t-1}, w_t) \nu_{t-1, m_{t-1}, w_t}.$$

If we extend the recursion by another step to define $\vec{\nu}_0$, we get that our minimization target is

$$\mathcal{L}_n(q_{(n)}; \theta_1) = \frac{1}{n} \operatorname*{E}_{M_0, W_1} \nu_{0, M_0, W_1}.$$

The minimization

$$V_t(\theta_t) = \min_{q_{(t,n)}} \operatorname*{E}_{M_{t-1}, W_t} \nu_{t-1, M_{t-1}, W_t}$$

can be looked at as the *cost-to-go* given the joint belief $\theta_t$ before step $t$. Importantly, the recursive formula 5, when minimized over $q_{(t,n)}$, is a Bellman equation. It contains a recursive term

$$\operatorname*{E}_{M_t, W_{t+1} | M_{t-1}, W_t} \nu_{t, M_t, W_{t+1}},$$

which is the expected future cost, and other terms which are the expected immediate costs, internal and external, of implementing $q_t$ in step $t$.

This suggests viewing our problem as a joint-belief MDP. Here the states are the joint beliefs $\theta_t$, the actions are $q_t$, and the next state always follows deterministically according to equation 3. This determinism allows us to use a time-dependent policy $q_{(n)}$, rather than a state-dependent one, and will prove useful in finding a solution.

The belief space of a standard POMDP can be looked at as the state space of a belief MDP, with the same actions and observations, and a linear transition function. If memory states are approximate beliefs, then our model is more like a further abstraction, where the MDP state space is the set of distributions over the belief space. Table 2 summarizes the main differences between this joint-belief MDP and the belief-MDP representation of discrete-action finite-horizon POMDPs.

---
[3]When the Lagrangian is written in terms of the policy and the initial joint belief, the other parameters are taken to be consistent with them.

Table 2. Differences in belief-MDP representation of POMDPs and Bounded Passive POMDPs

|  | POMDP | Bounded Passive POMDP |
| --- | --- | --- |
| State space | beliefs, $\Delta(\mathcal{W})$ | joint beliefs, $(\Delta(\mathcal{M}))^{\mathcal{W}}$ |
| Action space | same as POMDP discrete | memory-state transitions continuous |
| State transition | stochastic linear in belief | deterministic linear in joint belief |
| Policy cost | external cost linear in belief | internal+external cost $D_{KL}$+linear in joint belief |
| Value function | piecewise-linear concave in belief | continuous concave in joint belief |

One important difference is in the structure of the value function. The expected cost $\mathcal{L}_{n-t+1}$ of a fixed policy suffix $q_{(t,n)}$ consists of some linear terms of expected distortion, but also some strictly convex terms. The latter all take the form of a Kullback-Leibler divergence between $q_{t'}$, for some $t' \geq t$, and a marginal $\bar{q}_{t'}$, the latter depending on $\theta_t$ through equations 2 and the recursion 3.

That this cost is not linear makes the representation of the value function a challenge, but a greater difficulty is the size of the policy space, which is finite in discrete-action finite-horizon POMDPs, but continuous here. Minimizing over it does not yield a piecewise-linear function of the joint belief, although it is still continuous, and the convex mixing of policies shows that it is still concave[4]. It is unclear how to finitely represent the resulting value function in our case.

### 4.4. Bounded Planning Algorithm

Perhaps surprisingly, the determinism of the joint-belief MDP allows us to define a local criterion for optimality. Together with iterations of algorithm 2 which make local improvements, this will guarantee convergence to a local optimum.

Our algorithm is a simple forward-backward algorithm, with a building block (algorithm 2) which is itself forward-backward. In each iteration we compute recursively forward the joint beliefs $\theta_{(n)}$ for the current policy $q_{(n)}$. Then we compute recursively backward a new policy $q'_{(n)}$, by finding in each step $t$ a policy suffix which is locally optimal for $\theta_t$. The criterion for optimality is that in each step we can use either $q'_{(t+1,n)}$ from the previous step or $q_{(t+1,n)}$ from the previous iteration, and whichever leads to a lower cost is chosen.

**Theorem 5.** *Algorithm 3 converges monotonically to a limit cost $\mathcal{L}^*$. For any $\epsilon \geq 0$, any $q^r_{(n)}$ which costs within $\epsilon$ of $\mathcal{L}^*$ is also within $\epsilon$ of a local minimum of*

---
[4]If rewards are used instead of costs, the value function is convex.

**Algorithm 3** Passive POMDP Bounded Planning
**Input:** $\theta_1, p, \sigma, d, \gamma_C, \gamma_M, \gamma_S, n$
**Output:** locally optimal $q_{(n)}$
  $r \leftarrow 0$
  Initialize some suggestion for $q_{(n)}^r$
  **repeat**
    $\theta_1^r \leftarrow \theta_1$
    Compute $\theta_{(2,n)}^r$ from $\theta_1^r$ and $q_{(n-1)}^r$ (eq. 3)
    **for** $t \leftarrow n$ **to** 1 **do**
      $q_{(t,n)}^{r+1,t} \leftarrow \arg\min_{q_{(t,n)}} \mathcal{L}_{n-t+1}(q_{(t,n)}; \theta_t^r)$
      s.t. $q_{(t+1,n)} \in \left\{ q_{(t+1,n)}^{r+1,t+1}, q_{(t+1,n)}^r \right\}$ (alg. 2)
    **end for**
    $q_{(n)}^{r+1} \leftarrow q_{(n)}^{r+1,1}$
    $r \leftarrow r+1$
  **until** $\mathcal{L}_n(q_{(n)}^r; \theta_1)$ converges

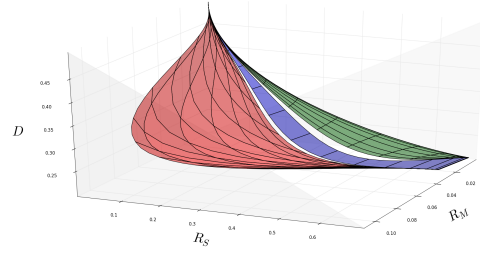

*Figure 2.* Boundary of the rate-distortion region for the sequential symmetric channel simulation
The parts from left to right: $\gamma_M = 0$; $\gamma_M = \gamma_S = 0$; $\gamma_S = 0$

*the bounded-inference-planning problem (section 2), in the sense that for any $1 \leq t \leq n$, the global minimum given $q_{(t-1)}^r$ and $q_{(t+1,n)}^r$ is at most $\epsilon$ better than $q_{(n)}^r$.*

*Proof.* In iteration $r$, $q_{(n)}^r$ from the previous iteration is feasible for $q_{(n)}^{r+1}$. Therefore the cost of $q_{(n)}^r$ is non-increasing in $r$ and converges monotonically to a limit $\mathcal{L}^*$.

Let $q_{(n)}^r$ be within some $\epsilon > 0$ of $\mathcal{L}^*$. Fix any $1 \leq t \leq n$, and let $q_t^*$ achieve the global optimum given $q_{(t-1)}^r$ and $q_{(t+1,n)}^r$. Then

$$\mathcal{L}_n(q_{(n)}^r; \theta_1) - \epsilon \leq \mathcal{L}_n(q_{(n)}^{r+1}; \theta_1)$$

$$\overset{(a)}{\leq} \mathcal{L}_n((q_{(t-1)}^r, q_{(t,n)}^{r+1,t}); \theta_1)$$

$$\overset{(b)}{\leq} \mathcal{L}_n((q_{(t-1)}^r, q_t^*, q_{(t+1,n)}^r); \theta_1),$$

where

(a) follows recursively from $(q_{t'}^r, q_{(t'+1,n)}^{r+1,t'+1})$ being feasible for $\theta_{t'}^r$ in iteration $r$, for each $1 \leq t' < t$, and

(b) follows from $(q_t^*, q_{(t+1,n)}^r)$ being feasible for $\theta_t^r$ in iteration $r$.

□

Where algorithm 3 runs algorithm 2, it can initialize $q_t$ to $q_t^r$ from the previous iteration. This may speed up each iteration, particularly when the algorithm has nearly converged. In addition, when running algorithm 3 with different sets of multipliers, it converges much faster if each run is initialized with the previous result. Empirically, this also leads to much better local minima if the runs are sorted in order of decreasing multipliers.

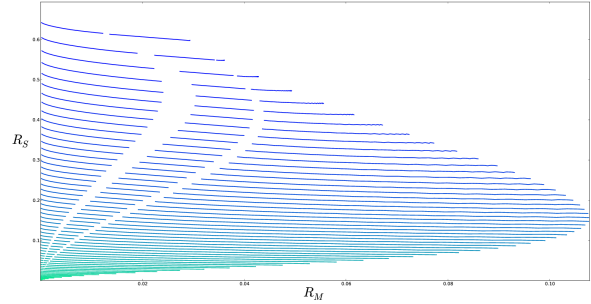

*Figure 3.* Contour map of the rate-distortion boundary for the sequential symmetric channel simulation

## 5. Simulations

### 5.1. Symmetric Channel

Figure 2 shows the boundary of the rate-distortion region for the 30-step sequential symmetric channel problem. The domains $\mathcal{W}$, $\mathcal{O}$ and $\mathcal{M}$ are all binary. The agent observes the state correctly with probability 0.8. The state remains the same for the next step independently with probability 0.8. The distortion is the delta function.

The boundary consists of three parts as in corollary 3. They have $\gamma_M = 0$ (left), $\gamma_M = \gamma_S = 0$ (middle) and $\gamma_S = 0$ (right). Empirically, taking $\gamma_C = 0$ is never feasible, as no optimal solution ever has $\mathcal{I}_C \leq \mathcal{I}_M + \mathcal{I}_S$.

To clarify this further, figure 3 shows a colored contour map of the boundary. The lower the distortion, the higher the required information rates. The tradeoff between memory and perception is illustrated by the negative slope of the contours.

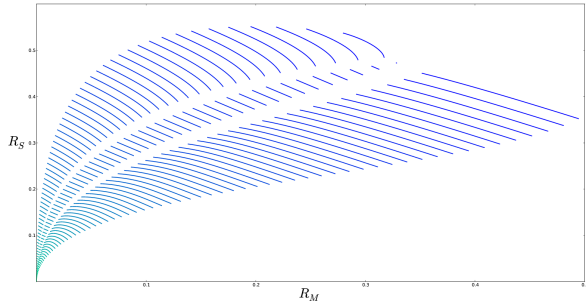

Figure 4. Contour map of the rate-distortion boundary for the Kelly gambling simulation

### 5.2. Kelly Gambling

Three horses are running in 10 races. Each horse has a fitness rating $f_i \in \{1, 2, 3\}$, and the winning horse is determined by softmax, i.e. horse $i$ wins with probability proportional to $\exp(f_i)$. Between the races, the fitness of each horse may independently grow by 1, with probability 0.1 if it is not maxed out, or drop by 1, with probability 0.1 if it is not depleted. Each horse keeps its fitness with the remaining probability.

The only observations are side races performed before each race: 2 random horses compete (with softmax) and the identities of the winner and the loser are announced. The memory state is a model of the world, consisting of the presumed fitness $\hat{f}_i$ of each horse. The log-optimal proportional gambling strategy is used (Kelly gambling, see Cover & Thomas, 2006), betting on horse $i$ a fraction of the wealth proportional to $\exp(\hat{f}_i)$. Each bet is double-or-nothing, and the distortion is the expected log return on the portfolio.

Figure 4 shows the contour map, which is not convex in this instance.

### 6. Conclusion

We have presented the problem of planning in Passive POMDPs with information-rate constraints. This problem takes the form of a sequential version of rate-distortion theory, and accordingly we were able to provide algorithms which globally optimize each step individually. Unfortunately, the full problem is not convex, and we expect that it has very hard instance sets.

Nevertheless, typical instances with some locality in their transitions and observations are expected to be easier. We have introduced an efficient and simple algorithm for finding a local minimum, and have used it to illustrate the problem with two simulations. In doing so, we have demonstrated the emergence of a memory-perception tradeoff in the problem.

Our work has been motivated by the problem of planning in general POMDPs, which may benefit from belief approximation which is principled by information theory. The application of our current results to this problem is left for future work.

### 7. Acknowledgement

This project is supported in part by the MSEE DARPA Project and by the Gatsby Charitable Foundation.